\documentclass[10pt,twocolumn,letterpaper]{article}

\usepackage{cvpr}              

\definecolor{cvprblue}{rgb}{0.21,0.49,0.74}
\usepackage{hyperref}

\def\paperID{*****} 
\def\confName{CVPR}
\def\confYear{2025}

\title{Enhancing Facial Expression Recognition through Dual-Direction Attention Mixed Feature Networks and CLIP: Application to 8th ABAW Challenge}

\author{Josep Cabacas-Maso$^1$, Elena Ortega-Beltrán$^1$, Ismael Benito-Altamirano$^{1,2}$, Carles Ventura$^1$
\\
\\
$^1$eHealth Center, Faculty of Computer Science, Multimedia and  Telecommunication, \\ Universitat Oberta de Catalunya, 08018 Barcelona, Spain\\
$^2$MIND/IN2UB, Department of Electronic and Biomedical Engineering, \\ Universitat de Barcelona, 08028 Barcelona, Spain\\
\{jcabacas, eortegabeltran, ibenitoal, cventuraroy\}@uoc.edu}

\begin{document}
\maketitle
\begin{abstract}
We present our contribution to the 8th ABAW challenge at CVPR 2025, where we tackle valence-arousal estimation, emotion recognition, and facial action unit detection as three independent challenges. Our approach leverages the well-known Dual-Direction Attention Mixed Feature Network (DDAMFN) for all three tasks, achieving results that surpass the proposed baselines. Additionally, we explore the use of CLIP for the emotion recognition challenge as an additional experiment. We provide insights into the architectural choices that contribute to the strong performance of our methods.
\end{abstract}    
\section{Introduction}
\label{sec:intro}

Facial emotion recognition has emerged as a pivotal area of research within affective computing, driven by its potential applications in fields ranging from human-computer interaction to psychological research and clinical diagnostics. Since Ekman's classification of human expression faces into emotions~\cite{ekman1978facial}, many studies have emerged in recent years. Calvo et al.~\cite{calvo2016faces}, Baltrusaities et al.~\cite{baltrusaitis2018openface}, or Kaya et al.~\cite{kaya2020exploring} laid the foundational framework for understanding facial expressions as a window into emotional states. In contemporary research, Liu et al.~\cite{liu2021deep} and Kim et al.~\cite{kim2023survey} continued to refine and expand these methodologies, by synthesizing insights from cognitive psychology, computer vision, and machine learning, researchers have made significant strides in enhancing the accuracy and applicability of facial emotion recognition systems. In addition, the integration of the dimensions of valence and arousal~\cite{soleymani2017deep,zafeiriou2017aff} added depth to the interpretation of emotional states, allowing more nuanced insight into human affective experiences.

Action unit detection~\cite{ekman2003darwin, lucey2010extended} complemented these efforts by parsing facial expressions into discrete muscle movements, facilitating a finer-grained analysis of emotional expressions across cultures and contexts. Such advancements not only improved the reliability of automated emotion recognition systems, but also opened the possibility to personalize affective computing applications in fields such as mental health monitoring~\cite{picard2020toward} or user experience design~\cite{zeng2019survey}.


To tackle all these challenges, researchers have explored innovative architectures such as the DDAMFN (Dual-Direction Attention Mixed Feature Network)~\cite{zhang2023ddamfn}. This novel approach integrates attention mechanisms~\cite{hou2021coordinate} and mixed feature extraction~\cite{chen2018mobilefacenets}, enhancing the network's ability to capture intricate details within facial expressions.

There is an increasing need to develop machines capable of understanding and appropriately responding to human emotions in real-world, day-to-day applications. Addressing this challenge, a series of competitions titled Affective Behavior Analysis in-the-Wild (ABAW) has been organized~\cite{kollias20247th,kollias20246th,kollias2024distribution,kollias2023abaw2,kollias2023multi,kollias2023abaw,kollias2022abaw,kollias2021analysing,kollias2021affect,kollias2021distribution,kollias2020analysing,kollias2019expression,kollias2019deep,kollias2019face,zafeiriou2017aff}. For the 8th ABAW ~\cite{Kollias2025,kolliasadvancements} challenge at CVPR 2025, six competitions were introduced: Valence-Arousal (VA) Estimation, Expression (EXPR) Recognition, Action Unit (AU) Detection, Compound Expression (CE) Recognition, Emotional Mimicry Intensity (EMI) Estimation, and Ambivalence/Hesitancy (AH) Recognition. In this work, we present our approach to the VA Estimation, EXPR Recognition, and AU Detection challenges, where we adapted the DDAMFN architecture for each challenge and additionally leveraged the CLIP embedding space to enhance emotion recognition.

\section{Methodology}

\subsection{Dataset curation}

For the 8th ABAW challenge, the organization provided the following datasets:
\begin{itemize}
    \item For VA Estimation: An augmented version of the Aff-Wild2 database~\cite{kollias20246th} will be used. This audiovisual (A/V), in-the-wild dataset consists of 594 videos, totaling approximately 3 million frames, with annotations for valence and arousal.
    \item For Expression Recognition: The Aff-Wild2 database will be used, comprising 548 videos with around 2.7 million frames. The dataset is annotated for the six basic expressions (anger, disgust, fear, happiness, sadness, and surprise), along with the neutral state and an other category, which represents affective states beyond the six basic emotions.
    \item For Action Unit Detection: The dataset includes 547 videos, also totaling approximately 2.7 million frames. It is annotated for 12 action units: AU1, AU2, AU4, AU6, AU7, AU10, AU12, AU15, AU23, AU24, AU25, and AU26.
\end{itemize}

As outlined in the dataset guidelines, we preprocessed the data by filtering out frames that contained annotation values outside the specified acceptable ranges. Specifically, any frames with valence/arousal values of -5, expression values of -1, or action unit (AU) values of -1 were excluded from consideration in our analysis. This process ensured that only frames with valid annotations were retained for model training and evaluation. The filtering criteria and the number of frames removed are summarized in ~\autoref{table:1}.

\begin{table}[h!]
  \centering
  \caption{Annotation ranges and invalid values.}
  \label{table:1}
  \begin{tabular}{@{}lcc@{}}
    \toprule
    \textbf{Annotation} & \textbf{Range} & \textbf{Invalid} \\ \midrule
    Valence/Arousal         & [0, 1]         & -5                      \\
    Expressions             & \{0, 1\}       & -1                      \\
    Action Units (AUs)      & \{0, 1\}       & -1                      \\ \bottomrule
  \end{tabular}
\end{table}

We applied rigorous filtering to the train-validation splits across three tasks: VA estimation, expression recognition, and AU detection. A summary of the dataset before and after filtering is shown in~\autoref{table:2}.

\begin{table}[h!]
  \centering
  \caption{Dataset summary after filtering}
  \label{table:2}
  \begin{tabular}{@{}lcc@{}}
    \toprule
    \textbf{VA estimation} & \textbf{Frames} & \textbf{Curated frames} \\ \midrule
    Training                   & 1,682,057                             & 1,679,854                                      \\
    Validation                 & 383,814                              & 382,021                                      \\
    Test                       & 927,210                              & -                                           \\ 
    \toprule
    \textbf{EXPR recognition} & \textbf{} & \textbf{} \\ \midrule
    Training                   & 1,110,367                             & 597,361                                      \\
    Validation                 & 453,535                              & 284,230                                      \\
    Test                       & 1,022,655                              & -                                           \\ 
    \toprule
    \textbf{AU detection} & \textbf{} & \textbf{} \\ \midrule
    Training                   & 1,381,411                             & 1,359,523                                      \\
    Validation                 & 454,022                              & 445,845                                      \\
    Test                       & 729,736                              & -                                           \\ \bottomrule
  \end{tabular}
\end{table}

\subsection{Network architecture}
\subsubsection{Dual-Direction Attention Mixed Feature Network}
For the 8th ABAW challenge, we adapted the Dual-Direction Attention Mixed Feature Network (DDAMFN)~\cite{zhang2023ddamfn} for three separate tasks. Each task has its own fully-connected layer at the end of the network: one for valence-arousal prediction with 2 output units, another for emotion recognition with 8 output units, and a third for action unit prediction with 12 output units.

\begin{figure*}[h!]
  \centering
  \includegraphics[width=0.85\textwidth]{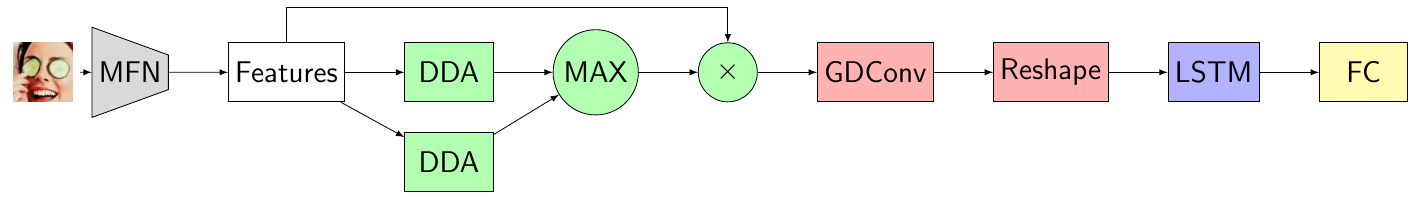}
  \caption{Our DDAMFN~\cite{zhang2023ddamfn} architecture for the 8th ABAW challenge: MobileFaceNet (MFN) for feature extraction (grey), Dual-Direction Attention (DDA) module (green), Global Depthwise Convolution (GDConv) layer (red), and one fully-connected layer (yellow) for the specific task being addressed (valence-arousal prediction, emotion recognition, and action unit detection).}
  \label{fig:cnn}
\end{figure*}

\autoref{fig:cnn} shows a diagram of the network, which features a base MobileFaceNet (MFN)~\cite{chen2018mobilefacenets} architecture for feature extraction, followed by a Dual-Direction Attention (DDA) module—with two attention heads—and a Global Depthwise Convolution (GDConv) layer. The output of the GDConv layer is reshaped and fed into the corresponding fully-connected layers for each individual task.

\subsubsection{Contrastive Language-Image Pretraining}
In addition to leveraging the Dual-Direction Attention Mixed Feature Network (DDAMFN) for the 8th ABAW challenge, we also employed the CLIP (Contrastive Language-Image Pretraining) model. CLIP \cite{radford2021learning} is a widely recognized Vision-Language model trained on millions of image-text pairs, where the text consists of image captions, using a contrastive learning approach. In essence, CLIP learns two types of embeddings: one for images and one for text. The model is trained to map matching image-text pairs to similar points in the embedding space, based on cosine similarity, while simultaneously pushing apart the embeddings of mismatched image-text pairs.

We followed the approach of \cite{10388075}, where they built on the CLIP embedding space by adding two fully connected layers on top of the image encoder to allow deformations of the embedding space. We then used a Contrastive loss with cosine similarity with a collection of text prompts. During training, both the CLIP image and text encoders are frozen, preserving the learned features while adapting the embedding space for sentiment analysis tasks.

\begin{figure*}[h!]
  \centering
  \includegraphics[width=0.85\textwidth]{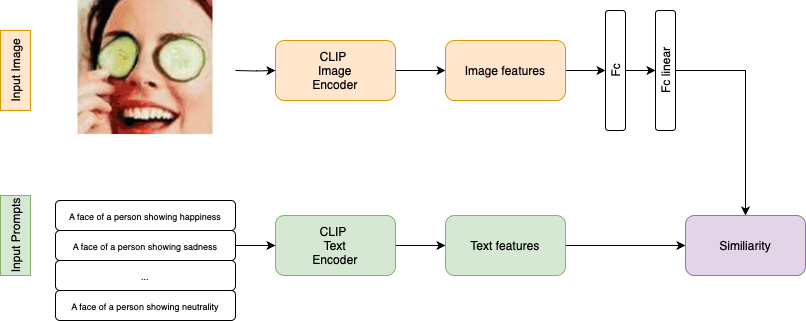}
  \caption{CLIP Architecture Overview: In this diagram, the architecture is divided into several key components. The visual path, highlighted in orange, processes the image input. The text path, shown in green, processes the textual input. The white areas represent the fully connected layers that bridge both paths. Finally, the similarity outputs, in purple, demonstrate how the model calculates the relationship between the image and text inputs.}
  \label{fig:CLIP}
\end{figure*}
\section{Training}
\subsection{DDAMFN}
Initially, the DDAMFN pretrained model was obtained from the stock versions available in the source code repository of the original work, which had been trained on the AffectNet-8 dataset~\cite{zhang2023ddamfn}. In our experiments, we preserved the pretrained weights for the feature extraction layers, attention mechanisms, and the GDConv layer, as these components had already been optimized for extracting meaningful features from the input data. However, the fully-connected layers were reinitialized with random weights to facilitate the adaptation of the model to our specific task.

To enhance the model’s performance across various challenges, we introduced a custom classifier for each individual task. These classifiers were designed to address the unique requirements of each challenge, allowing for more specialized learning. Each classifier was trained independently and separately, with the goal of adapting the pretrained feature extraction and attention components to the specific characteristics of the new tasks. This strategy ensured that the model could benefit from the strong generalization capabilities of the DDAMFN architecture while still tailoring its output to the specific demands of each challenge.

By training each custom classifier separately, we were able to achieve a fine-tuned balance between leveraging pretrained knowledge and optimizing the model for each individual task, leading to improved overall performance in the experiments.

Loss functions were calculated following these criteria:

\begin{itemize}
  \item For valence-arousal prediction, the loss function was calculated using the Concordance Correlation Coefficient (CCC). CCC is a measure that evaluates the agreement between two time series by assessing both their precision (how well the observations agree) and accuracy (how well the observations match the true values).
  \item For emotion recognition, was cross-entropy, which is commonly employed in classification tasks to measure the difference between the predicted probability distribution and the true distribution.
  \item For action unit (AU) detection, the binary cross-entropy loss was used, which is suitable for binary classification tasks, measuring the difference between the predicted probability and the actual binary outcome for each action unit.
\end{itemize}

Furthermore, for the Action Unit task, a global threshold of 0.5 was initially tested across all AUs, followed by individual optimization for each AU~\cite{savchenko2022framelevelpredictionfacialexpressions}.

However, after this initial round of experiments, we observed that the model could benefit from further improvements in capturing the temporal coherence between consecutive frames. To address this, we decided to replace the fully connected layers with a Long Short-Term Memory (LSTM) network. The LSTM was specifically chosen for its ability to retain and model long-term dependencies in sequential data, enabling the model to maintain temporal coherence across frames.

\subsection{CLIP Contrastive}
Following the work done in \cite{10388075} we used the same architecture and loss.
For the CLIP Contrastive model (Fig. \ref{fig:CLIP}), we introduced two fully connected layers on top of the image encoder to allow adjustments to the embedding space. We then used Contrastive loss with cosine similarity against a set of text prompts. During training, both the CLIP image and text encoders are kept frozen. A unique prompt was generated for each emotion based on the structure "a face showing \textit{emotion}."

In the CLIP Contrastive architecture, we added a fully connected layer with 512 units and ReLU activation on top of the CLIP image encoder, followed by another fully connected layer with 512 units and a linear activation function before the loss function. The linear activation function was selected to retain negative activations, as the original CLIP-trained embedding space contains both negative and positive values.

To train the model, we employed the same contrastive loss function used in the original CLIP approach \cite{radford2021learning}. Specifically, given a batch of \(N\) image-text pairs \((I_i, T_i), i = 1, \dots, N\), the losses for the image and text are computed as follows:

\begin{equation}
L_{\text{img}} = -\frac{1}{N} \sum_{i=1}^{N} \log \frac{\exp(\langle I_{ei}, T_{ei} \rangle)}{\sum_{j=1}^{N} \exp(\langle I_{ei}, T_{ej} \rangle)} \quad 
\end{equation}

\begin{equation}
L_{\text{text}} = -\frac{1}{N} \sum_{i=1}^{N} \log \frac{\exp(\langle T_{ei}, I_{ei} \rangle)}{\sum_{j=1}^{N} \exp(\langle T_{ei}, I_{ej} \rangle)} \quad
\end{equation}

where \(\langle I_{ei}, T_{ej} \rangle\) is the cosine similarity between the embedding vectors of the \(i\)-th image sample \(I_{ei}\) and the \(j\)-th text sample \(T_{ej}\), and \(\langle I_{ei}, T_{ei} \rangle\) is the cosine similarity between the \(i\)-th image sample and its corresponding text caption. The contrastive loss \(L_{\text{CO}}\) is computed as:

\begin{equation}
L_{\text{CO}} = \frac{L_{\text{img}} + L_{\text{text}}}{2} \quad 
\end{equation}

This loss drives the model to enhance the similarity between the embedding vector of each sample and its associated text description while reducing the similarity between the embedding vectors of each sample and the other samples in the batch. As we did previously in the experiments using DDAMFN, we have decided to replace the fully connected layers with an LSTM to maintain temporal coherence.

\section{Results}
The metrics evaluated for each challenge are the Concordance Correlation Coefficient (CCC) for valence, arousal, and their combination (valence-arousal), F1 score for emotion classification and F1 score for action unit (AU) detection:

\begin{equation}
  P = \frac{\textrm{CCC}_{V} + \textrm{CCC}_{A}}{2}
\end{equation}
\begin{equation}
  P =  \frac{\textrm{F1}_{Expr}}{8} 
\end{equation}
\begin{equation}
  P = \frac{\textrm{F1}_{AU}}{12}
\end{equation}

Table \ref{tab:headings-1} presents the performance metrics for each challenge across the different architectures tested. The metrics include the combined $\textrm{CCC}_{VA}$ (Valence-Arousal), $\textrm{F1}_{Expr}$ (expression classification), $\textrm{F1}_{AU}$ (action unit classification), and $\textrm{F1}_{AUopt}$ (optimized action unit classification after threshold optimization).

\begin{itemize}
    \item The baseline model shows the lowest performance across all tasks, with a $\textrm{CCC}_{VA}$ of 0.22, $\textrm{F1}_{Expr}$ of 0.25, and $\textrm{F1}_{AU}$ of 0.39. No $\textrm{F1}_{AUopt}$ value is reported for the baseline.
    \item The DDAMFN+Fc model improves upon the baseline, achieving a $\textrm{CCC}_{VA}$ of 0.416. However, the $\textrm{F1}_{Expr}$ and $\textrm{F1}_{AU}$ values (0.249 and 0.332, respectively) remain lower compared to other architectures. The $\textrm{F1}_{AUopt}$ score after threshold optimization is 0.362.
    \item The DDAMFN+LSTM model achieves the highest performance overall, with the highest $\textrm{CCC}_{VA}$ of 0.479, $\textrm{F1}_{AU}$ of 0.411, and $\textrm{F1}_{AUopt}$ of 0.451. These results indicate that this architecture performs best in terms of both emotional valence-arousal prediction and action unit classification.
    \item The CLIP+Fc and CLIP+LSTM models provided promising results in expression classification. Specifically, CLIP+LSTM achieved a $\textrm{F1}_{Expr}$ of 0.336, which is the highest among the rest of the architectures, demonstrating strong performance in this task. 
\end{itemize}

\begin{table}[ht]
  \caption{Performance metrics for each challenge. Table shows $\textrm{F1}_{Expr}$ and $\textrm{F1}_{AU}$ already normalized by the number of classes for each classification task. $\textrm{CCC}_{AV}$ is the combined CCC for valence and arousal. $\textrm{F1}_{AUopt}$ is the result after optimizing thresholds}
  \label{tab:headings-1}
  \centering
  \resizebox{0.45\textwidth}{!}{ 
    \begin{tabular}{@{}lcccccc@{}}
      \toprule
      Architecture &  $\textrm{CCC}_{VA}$ & $\textrm{F1}_{Expr}$ & $\textrm{F1}_{AU}$ & $\textrm{F1}_{AUopt}$ \\
      \midrule
      Baseline & \textbf{0.22} & \textbf{0.25} & \textbf{0.39} & - \\
      \midrule
      DDAMFN+Fc & 0.416 & 0.249 & 0.332 & 0.362 \\
      DDAMFN+LSTM & \textbf{0.479} & 0.289 & 0.411 & \textbf{0.451} \\
      CLIP+Fc  & - & 0.282 & -  & -\\
      CLIP+LSTM  & - & \textbf{0.341} & -  & -\\
      \bottomrule
    \end{tabular}
  }
\end{table}

\section{Conclusion}
The performance of the models can be attributed to several key architectural choices and design considerations. The DDAMFN+Fc model demonstrated a significant improvement over the baseline in emotional valence-arousal prediction, achieving a $\textrm{CCC}_{VA}$ of 0.416. However, its performance in expression classification ($\textrm{F1}_{Expr}$ = 0.249) and action unit classification ($\textrm{F1}_{AU}$ = 0.332) remained relatively lower compared to the other models, indicating that the model struggles with fine-grained expression and action unit recognition. This may be due to the limitations of the fully connected layers in capturing the complex spatial and temporal features required for these tasks. The architecture appears to perform better with simpler, less dynamic tasks like valence-arousal prediction, where temporal and spatial dependencies are less crucial.

In contrast, the DDAMFN+LSTM model performed the best overall, achieving the highest $\textrm{CCC}_{VA}$ (0.479), $\textrm{F1}_{AU}$ (0.411), and $\textrm{F1}_{AUopt}$ (0.451). The integration of LSTM layers in this model allowed it to capture the temporal dependencies inherent in facial expressions and emotional states, providing a distinct advantage for both action unit classification and valence-arousal prediction. The LSTM's ability to maintain memory of previous frames is crucial in tasks involving dynamic data, such as emotion recognition, where the progression of emotional states over time plays a critical role. This temporal awareness, when combined with DDAMFN’s feature extraction capabilities, enabled the DDAMFN+LSTM model to excel at both the prediction of emotional valence and arousal as well as the classification of facial action units, where subtle changes in facial expressions are key.

The CLIP+Fc and CLIP+LSTM models showed promising results, particularly in the domain of expression classification. The CLIP+LSTM model, with an $\textrm{F1}_{Expr}$ of 0.336, outperformed the other models in this task, demonstrating its strong ability to capture the nuances of facial expressions. This success can be attributed to the CLIP model’s capability to process both visual and contextual information, combined with the LSTM’s ability to capture temporal dependencies. This synergy enabled the model to recognize how facial expressions evolve over time, making it especially well-suited for the task of expression classification, where understanding transitions between facial states is crucial. Moreover, CLIP models, known for their robust performance in image-based tasks, likely enhanced the model’s ability to interpret diverse facial expressions and complex visual features, leading to a notable improvement in expression classification accuracy.

In summary, the superior performance of the DDAMFN+LSTM and CLIP+LSTM models can largely be attributed to their ability to handle both temporal and spatial features effectively. These models performed best in tasks that required an understanding of the dynamics of emotional valence-arousal and the subtlety of facial expressions. The LSTM’s strength in learning temporal patterns, combined with advanced feature extraction methods like those employed by DDAMFN and CLIP, provided these models with a significant advantage. This highlights the importance of incorporating temporal modeling and dynamic feature extraction in architectures designed for complex tasks such as emotion recognition and facial action unit classification.

{
    \small
    \bibliographystyle{ieeenat_fullname}
    \bibliography{main}

\begin{thebibliography}{34}
\providecommand{\natexlab}[1]{#1}
\providecommand{\url}[1]{\texttt{#1}}
\expandafter\ifx\csname urlstyle\endcsname\relax
  \providecommand{\doi}[1]{doi: #1}\else
  \providecommand{\doi}{doi: \begingroup \urlstyle{rm}\Url}\fi

\bibitem[Baltrusaitis et~al.(2018)Baltrusaitis, Zadeh, Lim, and Morency]{baltrusaitis2018openface}
Tadas Baltrusaitis, Amir Zadeh, Yagmur Lim, and Louis-Philippe Morency.
\newblock Openface 2.0: Facial behavior analysis toolkit.
\newblock \emph{in IEEE Winter Conference on Applications of Computer Vision (WACV)}, 2018.

\bibitem[Bustos et~al.(2023)Bustos, Civit, Du, Solé-Ribalta, and Lapedriza]{10388075}
Cristina Bustos, Carles Civit, Brian Du, Albert Solé-Ribalta, and Agata Lapedriza.
\newblock On the use of vision-language models for visual sentiment analysis: a study on clip.
\newblock In \emph{2023 11th International Conference on Affective Computing and Intelligent Interaction (ACII)}, pages 1--8, 2023.

\bibitem[Calvo and Nummenmaa(2016)]{calvo2016faces}
Manuel~G Calvo and Lauri Nummenmaa.
\newblock Faces and feelings: The universal recognition of emotions.
\newblock \emph{Annu. Rev. Psychol.}, 67:\penalty0 451--471, 2016.

\bibitem[Chen et~al.(2018)Chen, Liu, Gao, and Han]{chen2018mobilefacenets}
Sheng Chen, Yang Liu, Xiang Gao, and Zhen Han.
\newblock Mobilefacenets: Efficient cnns for accurate real-time face verification on mobile devices.
\newblock In \emph{Biometric Recognition: 13th Chinese Conference, CCBR 2018, Urumqi, China, August 11-12, 2018, Proceedings 13}, pages 428--438. Springer, 2018.

\bibitem[Ekman(2003)]{ekman2003darwin}
Paul Ekman.
\newblock Darwin and facial expression: A century of research in review.
\newblock \emph{General Psychology}, 7\penalty0 (2):\penalty0 121--125, 2003.

\bibitem[Ekman and Friesen(1978)]{ekman1978facial}
Paul Ekman and Wallace~V Friesen.
\newblock Facial action coding system: A technique for the measurement of facial movement.
\newblock \emph{Consulting Psychologists Press}, 1978.

\bibitem[Hou et~al.(2021)Hou, Zhou, and Feng]{hou2021coordinate}
Qibin Hou, Daquan Zhou, and Jiashi Feng.
\newblock Coordinate attention for efficient mobile network design.
\newblock In \emph{Proceedings of the IEEE/CVF conference on computer vision and pattern recognition}, pages 13713--13722, 2021.

\bibitem[Kaya and G{\"u}rpinar(2020)]{kaya2020exploring}
H{\"u}seyin Kaya and Emre G{\"u}rpinar.
\newblock Exploring deep convolutional neural networks for facial action unit recognition.
\newblock \emph{Neurocomputing}, 387:\penalty0 368--380, 2020.

\bibitem[Kim et~al.(2023)Kim, Lee, Kim, and Yoo]{kim2023survey}
Jihye Kim, Seung-Chan Lee, Jong-Soo Kim, and Chee~Sun Yoo.
\newblock A survey on facial emotion recognition: Approaches, databases, and challenges.
\newblock \emph{Pattern Recognition Letters}, 153:\penalty0 84--92, 2023.

\bibitem[Kollias(2022)]{kollias2022abaw}
Dimitrios Kollias.
\newblock Abaw: Valence-arousal estimation, expression recognition, action unit detection \& multi-task learning challenges.
\newblock In \emph{Proceedings of the IEEE/CVF Conference on Computer Vision and Pattern Recognition}, pages 2328--2336, 2022.

\bibitem[Kollias(2023{\natexlab{a}})]{kollias2023abaw}
Dimitrios Kollias.
\newblock Abaw: learning from synthetic data \& multi-task learning challenges.
\newblock In \emph{European Conference on Computer Vision}, pages 157--172. Springer, 2023{\natexlab{a}}.

\bibitem[Kollias(2023{\natexlab{b}})]{kollias2023multi}
Dimitrios Kollias.
\newblock Multi-label compound expression recognition: C-expr database \& network.
\newblock In \emph{Proceedings of the IEEE/CVF Conference on Computer Vision and Pattern Recognition}, pages 5589--5598, 2023{\natexlab{b}}.

\bibitem[Kollias and Zafeiriou(2019)]{kollias2019expression}
Dimitrios Kollias and Stefanos Zafeiriou.
\newblock Expression, affect, action unit recognition: Aff-wild2, multi-task learning and arcface.
\newblock \emph{arXiv preprint arXiv:1910.04855}, 2019.

\bibitem[Kollias and Zafeiriou(2021{\natexlab{a}})]{kollias2021affect}
Dimitrios Kollias and Stefanos Zafeiriou.
\newblock Affect analysis in-the-wild: Valence-arousal, expressions, action units and a unified framework.
\newblock \emph{arXiv preprint arXiv:2103.15792}, 2021{\natexlab{a}}.

\bibitem[Kollias and Zafeiriou(2021{\natexlab{b}})]{kollias2021analysing}
Dimitrios Kollias and Stefanos Zafeiriou.
\newblock Analysing affective behavior in the second abaw2 competition.
\newblock In \emph{Proceedings of the IEEE/CVF International Conference on Computer Vision}, pages 3652--3660, 2021{\natexlab{b}}.

\bibitem[Kollias et~al.({\natexlab{a}})Kollias, Schulc, Hajiyev, and Zafeiriou]{kollias2020analysing}
D Kollias, A Schulc, E Hajiyev, and S Zafeiriou.
\newblock Analysing affective behavior in the first abaw 2020 competition.
\newblock In \emph{2020 15th IEEE International Conference on Automatic Face and Gesture Recognition (FG 2020)(FG)}, pages 794--800, {\natexlab{a}}.

\bibitem[Kollias et~al.({\natexlab{b}})Kollias, Tzirakis, Cowen, Kotsia, Cogitat, Granger, Pedersoli, Bacon, Baird, Shao, et~al.]{kolliasadvancements}
Dimitrios Kollias, Panagiotis Tzirakis, Alan Cowen, Irene Kotsia, UK Cogitat, Eric Granger, Marco Pedersoli, Simon Bacon, Alice Baird, Chunchang Shao, et~al.
\newblock Advancements in affective and behavior analysis: The 8th abaw workshop and competition.
\newblock {\natexlab{b}}.

\bibitem[Kollias et~al.("2025")Kollias, Tzirakis, Cowen, Zafeiriou, Kotsia, Granger, Pedersoli, Bacon, Baird, Gagne, Shao, Hu, Belharbi, and Aslam]{Kollias2025}
Dimitrios Kollias, Panagiotis Tzirakis, Alan~S. Cowen, Stefanos Zafeiriou, Irene Kotsia, Eric Granger, Marco Pedersoli, Simon~L. Bacon, Alice Baird, Chris Gagne, Chunchang Shao, Guanyu Hu, Soufiane Belharbi, and Muhammad~Haseeb Aslam.
\newblock {Advancements in Affective and Behavior Analysis: The 8th ABAW Workshop and Competition}.
\newblock "2025".

\bibitem[Kollias et~al.(2019{\natexlab{a}})Kollias, Sharmanska, and Zafeiriou]{kollias2019face}
Dimitrios Kollias, Viktoriia Sharmanska, and Stefanos Zafeiriou.
\newblock Face behavior a la carte: Expressions, affect and action units in a single network.
\newblock \emph{arXiv preprint arXiv:1910.11111}, 2019{\natexlab{a}}.

\bibitem[Kollias et~al.(2019{\natexlab{b}})Kollias, Tzirakis, Nicolaou, Papaioannou, Zhao, Schuller, Kotsia, and Zafeiriou]{kollias2019deep}
Dimitrios Kollias, Panagiotis Tzirakis, Mihalis~A Nicolaou, Athanasios Papaioannou, Guoying Zhao, Bj{\"o}rn Schuller, Irene Kotsia, and Stefanos Zafeiriou.
\newblock Deep affect prediction in-the-wild: Aff-wild database and challenge, deep architectures, and beyond.
\newblock \emph{International Journal of Computer Vision}, pages 1--23, 2019{\natexlab{b}}.

\bibitem[Kollias et~al.(2021)Kollias, Sharmanska, and Zafeiriou]{kollias2021distribution}
Dimitrios Kollias, Viktoriia Sharmanska, and Stefanos Zafeiriou.
\newblock Distribution matching for heterogeneous multi-task learning: a large-scale face study.
\newblock \emph{arXiv preprint arXiv:2105.03790}, 2021.

\bibitem[Kollias et~al.(2023)Kollias, Tzirakis, Baird, Cowen, and Zafeiriou]{kollias2023abaw2}
Dimitrios Kollias, Panagiotis Tzirakis, Alice Baird, Alan Cowen, and Stefanos Zafeiriou.
\newblock Abaw: Valence-arousal estimation, expression recognition, action unit detection \& emotional reaction intensity estimation challenges.
\newblock In \emph{Proceedings of the IEEE/CVF Conference on Computer Vision and Pattern Recognition}, pages 5888--5897, 2023.

\bibitem[Kollias et~al.(2024{\natexlab{a}})Kollias, Sharmanska, and Zafeiriou]{kollias2024distribution}
Dimitrios Kollias, Viktoriia Sharmanska, and Stefanos Zafeiriou.
\newblock Distribution matching for multi-task learning of classification tasks: a large-scale study on faces \& beyond.
\newblock In \emph{Proceedings of the AAAI Conference on Artificial Intelligence}, pages 2813--2821, 2024{\natexlab{a}}.

\bibitem[Kollias et~al.(2024{\natexlab{b}})Kollias, Tzirakis, Cowen, Zafeiriou, Kotsia, Baird, Gagne, Shao, and Hu]{kollias20246th}
Dimitrios Kollias, Panagiotis Tzirakis, Alan Cowen, Stefanos Zafeiriou, Irene Kotsia, Alice Baird, Chris Gagne, Chunchang Shao, and Guanyu Hu.
\newblock The 6th affective behavior analysis in-the-wild (abaw) competition.
\newblock In \emph{Proceedings of the IEEE/CVF Conference on Computer Vision and Pattern Recognition}, pages 4587--4598, 2024{\natexlab{b}}.

\bibitem[Kollias et~al.(2024{\natexlab{c}})Kollias, Zafeiriou, Kotsia, Dhall, Ghosh, Shao, and Hu]{kollias20247th}
Dimitrios Kollias, Stefanos Zafeiriou, Irene Kotsia, Abhinav Dhall, Shreya Ghosh, Chunchang Shao, and Guanyu Hu.
\newblock 7th abaw competition: Multi-task learning and compound expression recognition.
\newblock \emph{arXiv preprint arXiv:2407.03835}, 2024{\natexlab{c}}.

\bibitem[Liu et~al.(2021)Liu, Song, and Tan]{liu2021deep}
Xiaoying Liu, Xianhua Song, and Zheng-Hua Tan.
\newblock Deep learning for emotion recognition: A comprehensive review.
\newblock \emph{Neurocomputing}, 440:\penalty0 167--180, 2021.

\bibitem[Lucey et~al.(2010)Lucey, Cohn, Kanade, Saragih, Ambadar, and Matthews]{lucey2010extended}
Patrick Lucey, Jeffrey~F Cohn, Takeo Kanade, Jason Saragih, Zara Ambadar, and Iain Matthews.
\newblock The extended cohn-kanade dataset (ck+): A complete dataset for action unit and emotion-specified expression.
\newblock \emph{in IEEE Conference on Computer Vision and Pattern Recognition Workshops (CVPRW)}, 2010.

\bibitem[Picard et~al.(2020)Picard, Vyzas, and Healey]{picard2020toward}
Rosalind~W Picard, Elias Vyzas, and Jennifer Healey.
\newblock Toward machine emotional intelligence: Analysis of affective physiological states.
\newblock \emph{IEEE Transactions on Pattern Analysis and Machine Intelligence}, 23\penalty0 (10):\penalty0 1175--1191, 2020.

\bibitem[Radford et~al.(2021)Radford, Kim, Hallacy, Ramesh, Goh, Agarwal, Sastry, Askell, Mishkin, Clark, et~al.]{radford2021learning}
A. Radford, J.~W. Kim, C. Hallacy, A. Ramesh, G. Goh, S. Agarwal, G. Sastry, A. Askell, P. Mishkin, J. Clark, et~al.
\newblock Learning transferable visual models from natural language supervision.
\newblock In \emph{Proceedings of the 38th International Conference on Machine Learning}, pages 8748--8763. PMLR, 2021.

\bibitem[Savchenko(2022)]{savchenko2022framelevelpredictionfacialexpressions}
Andrey~V. Savchenko.
\newblock Frame-level prediction of facial expressions, valence, arousal and action units for mobile devices, 2022.

\bibitem[Soleymani et~al.(2017)Soleymani, Garcia, Jou, and Schuller]{soleymani2017deep}
Mohammad Soleymani, David Garcia, Brendan Jou, and Bj{\"o}rn Schuller.
\newblock Deep learning for affective computing: Text-based emotion recognition in decision support.
\newblock \emph{IEEE Transactions on Affective Computing}, 8\penalty0 (1):\penalty0 17--37, 2017.

\bibitem[Zafeiriou et~al.(2017)Zafeiriou, Kollias, Nicolaou, Papaioannou, Zhao, and Kotsia]{zafeiriou2017aff}
Stefanos Zafeiriou, Dimitrios Kollias, Mihalis~A Nicolaou, Athanasios Papaioannou, Guoying Zhao, and Irene Kotsia.
\newblock Aff-wild: Valence and arousal ‘in-the-wild’challenge.
\newblock In \emph{Computer Vision and Pattern Recognition Workshops (CVPRW), 2017 IEEE Conference on}, pages 1980--1987. IEEE, 2017.

\bibitem[Zeng et~al.(2019)Zeng, Pantic, Roisman, and Huang]{zeng2019survey}
Zhihong Zeng, Maja Pantic, Glenn~I Roisman, and Thomas~S Huang.
\newblock A survey of affect recognition methods: Audio, visual, and spontaneous expressions.
\newblock \emph{IEEE Transactions on Pattern Analysis and Machine Intelligence}, 31\penalty0 (1):\penalty0 39--58, 2019.

\bibitem[Zhang et~al.(2023)Zhang, Zhang, Zhang, Wang, and Song]{zhang2023ddamfn}
Saining Zhang, Yuhang Zhang, Ye Zhang, Yufei Wang, and Zhigang Song.
\newblock A dual-direction attention mixed feature network for facial expression recognition.
\newblock \emph{Electronics}, 12\penalty0 (17):\penalty0 3595, 2023.

\end{thebibliography}
}


\end{document}